\documentclass{article}

\PassOptionsToPackage{table}{xcolor}
\usepackage[margin=1in]{geometry}
\usepackage{times}
\usepackage[round,authoryear]{natbib}

\usepackage{amsmath,amssymb}
\usepackage{booktabs}
\usepackage{graphicx}
\usepackage{multirow}
\usepackage{makecell}
\usepackage{microtype}
\usepackage{xcolor}
\usepackage{float}
\usepackage{placeins}
\usepackage{pgfplots}
\usepackage{pgfplotstable}
\usepgfplotslibrary{groupplots}
\pgfplotsset{compat=1.18}
\usepackage{hyperref}
\usepackage{url}
\usepackage{dblfloatfix}

\definecolor{pixelnoise}{HTML}{0072B2}
\definecolor{framenoise}{HTML}{D55E00}

\pgfplotstableread{
sigma routepixel routeframe dirpixel dirframe distpixel distframe roompixel roomframe
0.00 53.61 53.61 73.45 73.45 49.31 49.31 64.96 64.96
0.02 51.55 51.03 73.24 73.24 48.98 49.77 53.76 54.86
0.05 48.97 47.94 72.52 72.62 49.52 49.59 48.86 51.83
0.10 50.00 47.42 72.62 71.28 48.84 49.48 44.07 50.63
0.20 46.39 51.55 71.49 69.94 48.64 49.57 41.70 50.44
}\depthnoisedata

\graphicspath{{figures/}}

\title{GeoLift: Grounding Video Spatial Reasoning in Explicit 3D Maps}
\author{\begin{tabular}{c}
Xinglin Li$^{1,*}$ \quad TingTing Long$^{2,*}$ \quad Jingzhi Zhou$^{1}$ \\
Jiajing Chen$^{3}$ \quad Jian Yang$^{1}$ \quad Jin Xie$^{4,\dagger}$ \\
{\normalfont $^{1}$Nankai University \quad $^{2}$Peking University} \\
{\normalfont $^{3}$Amazon \quad $^{4}$Nanjing University} \\
{\normalfont\small $^{*}$Equal contribution. \quad $^{\dagger}$Corresponding author.}
\end{tabular}}

\date{}
\hypersetup{pdftitle={GeoLift: Grounding Video Spatial Reasoning in Explicit 3D Maps},pdfauthor={Xinglin Li, TingTing Long, Jingzhi Zhou, Jiajing Chen, Jian Yang, Jin Xie}}

\begin{document}

\maketitle

\begin{abstract} 

Video spatial reasoning requires multimodal large language models (MLLMs) to understand how objects are arranged and related across viewpoints. While these models can recognize objects in individual frames, integrating these observations to solve complex spatial tasks remains challenging. We examine route planning as a representative task and observe that existing models become substantially less accurate as the number of required turns increases, with several models achieving accuracy close to random guessing on routes requiring three or more turns. This finding suggests that recognizing landmarks alone is insufficient: models must also connect observations across viewpoints and track how landmarks are arranged in a shared space. Motivated by this observation, we introduce GeoLift, a training-free framework that uses feed-forward 3D reconstruction to map objects detected in individual frames into a shared metric 3D scene. GeoLift combines a frozen MLLM for 2D grounding with pretrained geometry models for scene reconstruction. It then uses shared geometric regions, called superpoints, to associate observations of the same object across views without a learned matcher. The reconstructed 3D map enables explicit computation of spatial relations and multistep routes, and provides geometric evidence for estimating distances and sizes. With all model weights frozen, GeoLift improves Qwen3-VL-4B's average score on VSI-Bench from 55.4\% to 64.1\%, including an increase in route-planning accuracy from 30.9\% to 53.6\%, and achieves 59.8\% accuracy on MindCube. These results show that connecting frame-level object grounding through a shared 3D map can improve complex spatial reasoning without task-specific training.

\end{abstract}

\section{Introduction}

Spatial reasoning requires more than recognizing what appears in an image. An agent must understand where objects are, how they relate to one another, and how these relationships change with the observer's viewpoint. These abilities are essential for navigation, manipulation, and interaction with the physical environment. Human spatial cognition integrates observations from different viewpoints into stable representations of places and spatial relations~\citep{orban2011extraction,kravitz2011new,epstein1998cortical,maguire1998knowing}. Developing similar capabilities in multimodal large language models (MLLMs) is a central challenge in video and multiview spatial reasoning.

A video presents a sequence of partial views rather than a complete description of a scene. As the camera moves, the same object appears at different image locations, while objects observed in separate frames may never be visible together. Answering spatial questions therefore requires connecting observations across viewpoints and understanding their relationships in a shared space. Modern MLLMs can recognize and ground objects in individual frames, but these capabilities do not necessarily translate into reliable scene-level spatial reasoning~\citep{yin2025mindcube}.

\begin{figure*}[t]
    \centering
    \includegraphics[width=\textwidth]{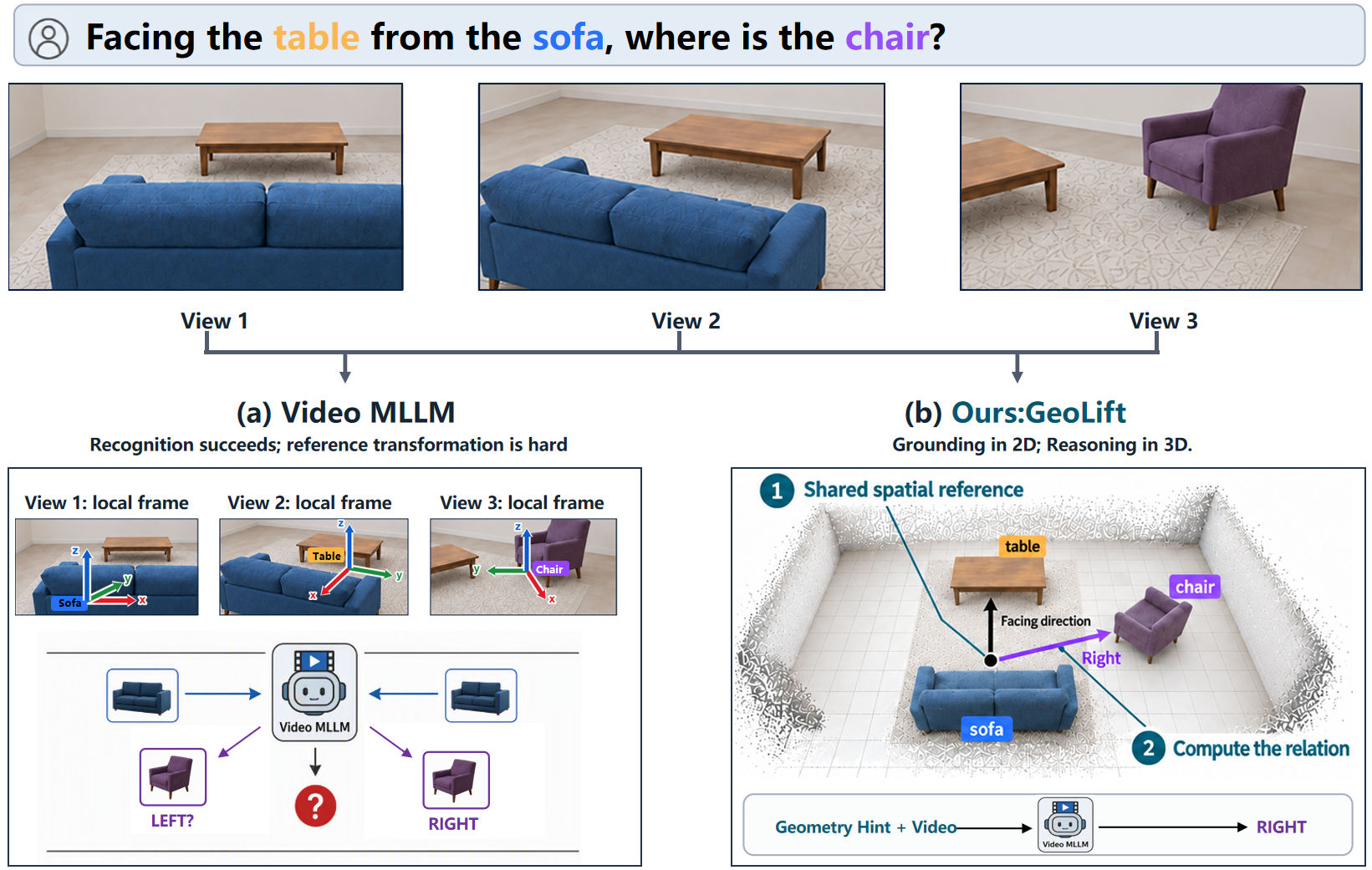}
    \caption{\textbf{From local 2D observations to explicit 3D spatial reasoning.}
    Given the same RGB video and spatial query, \textbf{(a) Video MLLM}
    grounds objects in individual frames, but each observation remains tied to
    its local coordinate frame. Without a shared reference, the cross-view
    relation between the sofa and chair is ambiguous. \textbf{(b) GeoLift}
    lifts the grounded objects into a shared spatial reference and explicitly
    computes the relation. For this query, the sofa-to-table direction defines
    the facing orientation, which places the chair on the right.}
    \label{fig:highlight}
\end{figure*}

Existing approaches often improve spatial reasoning by training MLLMs to incorporate geometric information, either through feature fusion or geometric supervision  \citep{Zheng2025LearningFV,Wu2025SpatialMLLMBM,Fan2026VLM3R,Yeh2026GASP}. As illustrated in Figure~\ref{fig:highlight} (a), representative geometry-augmented methods fuse visual and geometric features and infer spatial answers from learned representations. However, our route-planning analysis reveals that several direct and spatially trained MLLMs perform close to random guessing as the number of required turns increases (section ~\ref{motivation}). This finding suggests that geometric awareness alone may not be sufficient for complex tasks: models also need to maintain a consistent spatial reference and compose relations across multiple observations.

Our key insight is that feed-forward 3D reconstruction can provide this shared reference by mapping frame-level 2D groundings into a common 3D space. As illustrated in Figure~\ref{fig:highlight} (b), grounding objects in an explicit 3D map allows their spatial relationships to be computed directly. Building on this idea, we introduce GeoLift, a training-free framework that combines frozen MLLM grounding with reconstructed metric geometry. GeoLift uses geometric superpoints~\citep{landrieu2018large, robert2023efficient} to associate observations of the same object across viewpoints, producing a view-consistent map for explicit spatial computation and geometry-guided reasoning. All neural components remain frozen, with no benchmark-specific parameter updates.

We evaluate GeoLift on VSI-Bench and MindCube, covering egocentric videos and sparse multiview observations. With Qwen3-VL-4B, GeoLift improves the average VSI-Bench score from 55.4\% to 64.1\%, including an increase in route-planning accuracy from 30.9\% to 53.6\%, and achieves 59.8\% overall accuracy on MindCube. These results support the value of connecting local visual groundings through a shared 3D map, particularly for tasks that require combining spatial relations across viewpoints.
Our contributions are:
\begin{itemize}
\item We identify a weakness in existing MLLMs on route-planning tasks requiring multiple turns, motivating a shared spatial reference for connecting observations across viewpoints and composing spatial relations.
\item We introduce GeoLift, a training-free framework that lifts 2D groundings into a reconstructed metric 3D map and associates observations across viewpoints through geometric superpoint agreement, enabling explicit spatial computation and geometry-guided reasoning.
\item With all model weights frozen, GeoLift improves Qwen3-VL-4B's average score on VSI-Bench from 55.4\% to 64.1\%, with gains of 28.3 and 22.7 percentage points in relative-direction and route-planning accuracy, respectively. On MindCube, it improves overall accuracy from 23.4\% to 59.8\%, a gain of 36.4 percentage points.
\end{itemize}

\section{Related Work}

\subsection{Video and Multiview Spatial Reasoning Benchmarks}

Video and multiview spatial reasoning requires models to integrate evidence distributed across frames and viewpoints, especially when no single observation contains the full spatial context. VSI-Bench~\citep{2024Thinking} evaluates egocentric video reasoning through object counting, relative and absolute distance, relative direction, object and room size, route planning, and appearance order. MindCube~\citep{yin2025mindcube} complements this setting by studying cognitive mapping, perspective taking, and counterfactual spatial inference from limited multiview observations. Together, these benchmarks test whether a model can maintain a coherent spatial state across observations for metric comparison, viewpoint transformation, object relations, and multistep reasoning, revealing the gap between strong frame-level recognition and reliable scene-level spatial intelligence.

\subsection{Learning-Based Spatial Reasoning}

Learning-based approaches improve spatial reasoning through task-specific spatial supervision, post-training, or learned geometric interfaces. Video-R1, Cambrian-S, and InternSpatial advance this direction through spatially annotated data and supervised or reinforcement learning~\citep{Feng2025VideoR1,Yang2026CambrianS,Deng2026InternSpatial}. Geometry-augmented methods, including VG-LLM, Spatial-MLLM, and VLM-3R, align depth, camera, or multiview geometric features with visual-language representations through learned fusion~\citep{Zheng2025LearningFV,Wu2025SpatialMLLMBM,Fan2026VLM3R}. Structured representations provide another route: LASAR and EagleVision introduce latent cognitive maps and BEV representations, respectively~\citep{Tang2026LASAR,Wan2026EagleVision}, while VoxAnchor explicitly aligns cross-view visual features in a voxel-semantic space~\citep{Li2026VoxAnchor}. These approaches demonstrate the value of spatially organized representations, but generally require additional learning to align geometric cues with MLLM reasoning. Their spatial states therefore remain latent or are accessed through a learned query interface, motivating a complementary design based on explicit geometric computation.

\subsection{Training-Free Spatial Reasoning}

Training-free methods improve spatial reasoning at inference time by composing frozen MLLMs with prompting, rendering, or pretrained perception modules. VideoAnchor reinforces consistent cross-frame visual cues~\citep{Wang2026VideoAnchor}, TRACE organizes spatial evidence into textual representations~\citep{Hua2026TRACE}, and Reason, then Re-reason uses rendered novel views to verify and revise spatial hypotheses~\citep{Ma2026ReReason}. Other approaches construct explicit spatial representations using pretrained perception models. GeoLift combines frozen semantic and geometric models with geometric superpoints, lifting 2D groundings into a fused point cloud and associating cross-view observations without a learned matcher. No components are updated and no benchmark-specific spatial-QA supervision is used. Unlike text, latent-feature, or rendered-view interfaces, GeoLift uses a shared metric 3D workspace for explicit spatial computation.

\section{Method}

\subsection{Problem Setup and Motivation}
\label{motivation}

\paragraph{Problem setup.}
Let \(V=\{I_t\}_{t=1}^{T}\) denote an RGB video or a sparse collection of views, let \(q\) be a spatial query, and let \(y\) be its answer. The goal is to predict \(y\) from \((V,q)\) without updating model parameters on the target benchmark. The central challenge is to transform partial observations from changing viewpoints into a spatial representation that supports consistent reasoning.

\begin{figure*}[t]
    \centering
    \includegraphics[width=\textwidth]{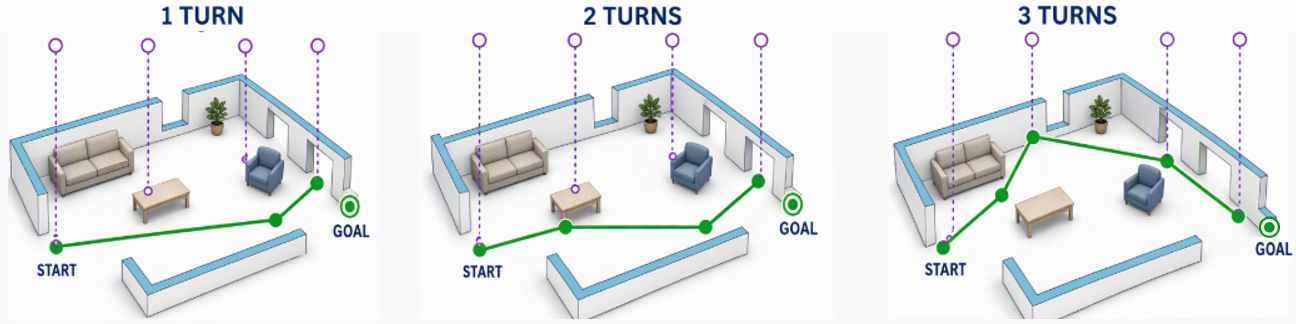}
    \caption{\textbf{Route-planning complexity as the number of turns increases.}
    The green path connects the start and goal through an ordered sequence of
    intermediate landmarks. A higher turn count requires more direction changes
    and stronger consistency across observations from different viewpoints,
    making the underlying global spatial arrangement increasingly important.}
    \label{fig:turns}
\end{figure*}

\begin{table*}[!b]
    \centering
    \small
    \setlength{\tabcolsep}{4.8pt}
    \caption{\textbf{Longer routes expose the need for a persistent spatial state.}
    Route-planning accuracy is grouped by the number of required turns.
    Values are accuracies (\%), with differences from the corresponding
    random-guessing baseline shown in parentheses in percentage points.
    Red entries are below the baseline, while orange entries are at most
    five percentage points above it.}
    \label{tab:route_motivation}
    \begin{tabular}{@{}l l c c c@{}}
        \toprule
        \textbf{Model} & \textbf{Protocol}
        & \textbf{1 turn} & \textbf{2 turns}
        & \textbf{$\geq$3 turns} \\
        & & \scriptsize 3 choices & \scriptsize 4 choices
        & \scriptsize 4 choices \\
        \midrule

        Qwen3-VL-4B & frozen
        & 43.4 {\scriptsize ($+10.1$)}
        & \textcolor{red!75!black}{22.5 {\scriptsize ($-2.5$)}}
        & \textcolor{red!75!black}{23.7 {\scriptsize ($-1.3$)}} \\

        GPT-5.6-Sol & frozen
        & 48.7 {\scriptsize ($+15.4$)}
        & 36.3 {\scriptsize ($+11.3$)}
        & \textcolor{orange!80!black}{28.9 {\scriptsize ($+3.9$)}} \\

        EgoMind 7B & learned
        & \textcolor{orange!80!black}{34.2 {\scriptsize ($+0.9$)}}
        & \textcolor{orange!80!black}{28.8 {\scriptsize ($+3.8$)}}
        & \textcolor{orange!80!black}{28.9 {\scriptsize ($+3.9$)}} \\

        Cambrian-S 7B (32f) & learned
        & 44.7 {\scriptsize ($+11.4$)}
        & 41.3 {\scriptsize ($+16.3$)}
        & \textcolor{orange!80!black}{26.3 {\scriptsize ($+1.3$)}} \\

        Cambrian-S 7B (64f) & learned
        & 53.9 {\scriptsize ($+20.6$)}
        & 35.0 {\scriptsize ($+10.0$)}
        & 31.6 {\scriptsize ($+6.6$)} \\

        GS-Reasoner & learned
        & 55.3 {\scriptsize ($+22.0$)}
        & \textcolor{orange!80!black}{30.0 {\scriptsize ($+5.0$)}}
        & \textcolor{orange!80!black}{26.3 {\scriptsize ($+1.3$)}} \\

        VLM-3R 7B & learned
        & 50.0 {\scriptsize ($+16.7$)}
        & 45.0 {\scriptsize ($+20.0$)}
        & 39.5 {\scriptsize ($+14.5$)} \\

        \textbf{GeoLift (ours)} & \textbf{frozen}
        & \textbf{57.9} {\scriptsize ($+24.6$)}
        & \textbf{52.5} {\scriptsize ($+27.5$)}
        & \textbf{47.4} {\scriptsize ($+22.4$)} \\

        \midrule
        Random guessing & --
        & 33.3 & 25.0 & 25.0 \\

        \bottomrule
    \end{tabular}

    \vspace{1mm}
    \scriptsize
    The three groups contain 76, 80, and 38 questions, respectively;
    the $\geq$3-turn group combines questions requiring 3 and 4 turns.
    Differences are computed from the displayed accuracies and baselines.
\end{table*}

\paragraph{Motivation.} 
We observe that current video MLLMs remain weak on complex spatial reasoning tasks that require combining information across viewpoints and reasoning over multiple steps. In this section, we examine this limitation through route planning, where a model must connect landmarks observed in different frames and determine an ordered sequence of turns. To study how performance changes with reasoning complexity, we group route-planning questions from VSI-Bench by the number of required turns: one, two, and three or more, as examples shown on Figure~\ref{fig:turns}. We compare direct MLLM inference, spatially trained models, and GeoLift within each group. Because one-turn questions have three answer choices while multi-turn questions have four, we also report the corresponding random-choice accuracy of 33.3\% and 25.0\%, respectively.

Table~\ref{tab:route_motivation} shows that existing models generally become less accurate as more turns are required. On questions with three or more turns, Qwen3-VL-4B achieves 23.7\% accuracy, while GS-Reasoner and Cambrian-S with 32 frames both achieve 26.3\%, close to random guessing. These results suggest that recognizing individual landmarks and incorporating geometric cues do not by themselves ensure reliable multistep spatial reasoning. A model must also associate the same landmark across viewpoints, maintain its position relative to other landmarks, and update the reference direction after each turn. Errors in these operations may accumulate across successive steps, making routes with more turns particularly challenging.

GeoLift addresses these requirements by organizing frame-level 2D groundings in a shared reconstructed 3D map, where landmark observations can be associated and spatial relations explicitly computed and composed. On questions requiring three or more turns, GeoLift achieves 47.4\% accuracy, exceeding the direct Qwen3-VL-4B baseline by 23.7 percentage points. Although this task remains challenging, the improvement supports the value of a shared geometric reference for multistep spatial reasoning. The following sections describe how GeoLift constructs and uses this reference.

\subsection{GeoLift Overview}
\begin{figure*}[t]
    \centering
    \includegraphics[width=\textwidth]{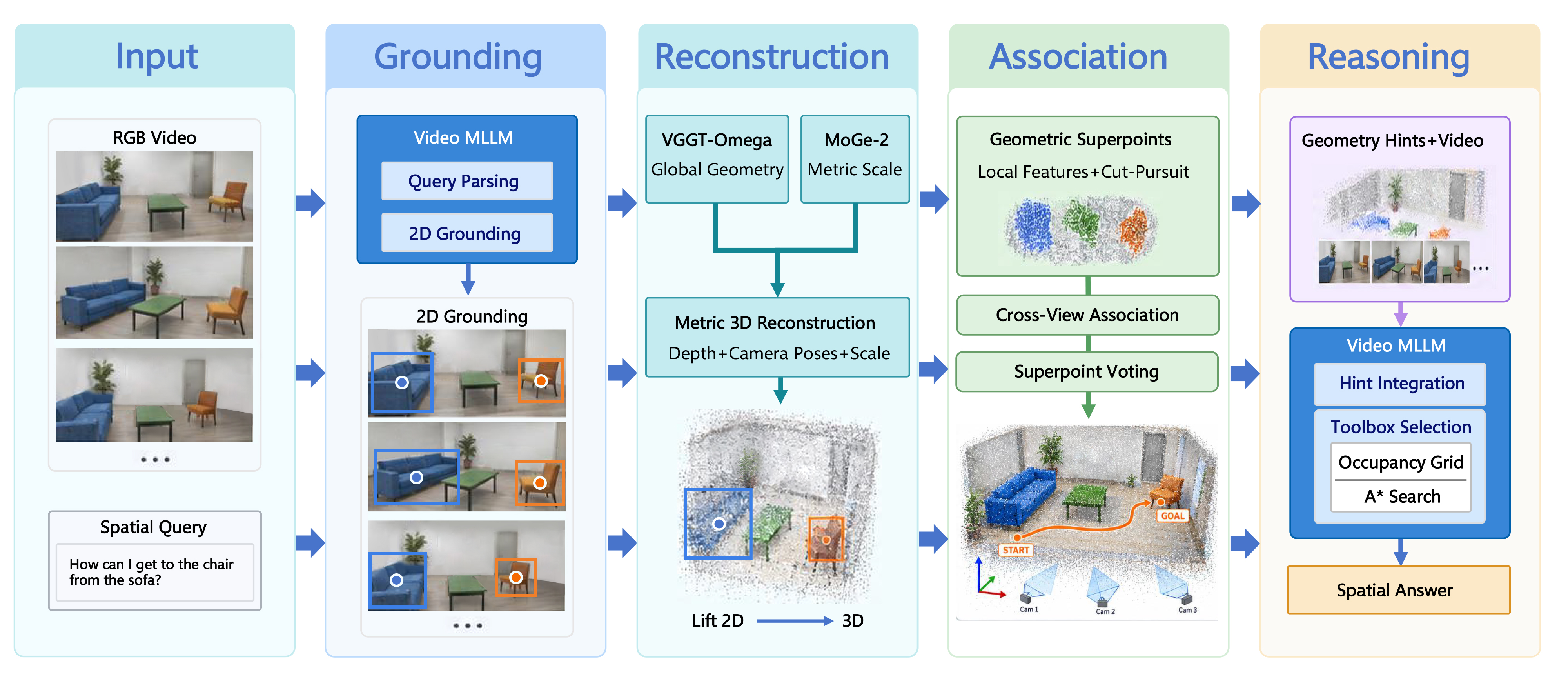}
    \caption{\textbf{GeoLift framework.}
    A frozen Video MLLM parses the query and grounds query-relevant entities
    in the RGB video. VGGT-Omega and MoGe-2 recover global geometry and metric
    scale, allowing the 2D groundings to be lifted into a shared 3D reference.
    Geometric superpoints and superpoint agreement associate observations of
    the same instance across views without a learned matcher. The resulting
    view-consistent map supports a toolbox of deterministic spatial operations,
    including occupancy-grid construction and A* route search. Geometry can also be provided as a hint together
    with the video for queries that require visual interpretation.}
    \label{fig:framework}
\end{figure*}

Figure~\ref{fig:framework} illustrates the pipeline of GeoLift, which consists of four stages: grounding, reconstruction, association, and reasoning. Given an RGB video and a spatial query, a frozen video MLLM first identifies and localizes query-relevant objects in individual frames. VGGT-Omega reconstructs the scene geometry, while MoGe-2 calibrates its metric scale, allowing these 2D groundings to be mapped into a shared 3D space. GeoLift then uses geometric superpoint agreement to associate observations of the same object across viewpoints, producing a 3D map with consistent object identities. Finally, the reconstructed map supports spatial reasoning in two ways: directly computing spatial relations and routes, or providing geometric evidence alongside the original video to guide the MLLM's answer. All neural components remain frozen throughout the pipeline. The following sections describe each stage in detail.

\subsection{Grounding}
The grounding module localizes query-relevant objects in individual frames for subsequent mapping into 3D. A frozen video MLLM first searches a grid of sampled frames with frame indices, then refines the target bounding boxes in the selected original frames. Missing targets trigger a denser frame search. For sparse multiview inputs, objects are localized independently in each view, with separate prompts for individual objects when needed. Each observation records the object name, frame or view index, normalized 2D bounding box, and confidence score, with an optional 2D point for navigation landmarks. Low-confidence observations and invalid boxes are discarded before 3D lifting and cross-view association.

\subsection{Reconstruction}
The reconstruction module builds a shared metric 3D space and maps the grounded 2D observations into it. It consists of two steps: reconstructing the scene geometry and lifting object observations from individual frames into 3D.

\paragraph{3D reconstruction.}
We use frozen VGGT-Omega~\citep{wang2026vggt} to predict a depth map $d_f^{\mathrm{V}}$, camera intrinsics $K_f$, and a camera-to-world pose $[R_f \mid t_f]$ for each frame $f$. These predictions place observations from different views in a common coordinate system, but the reconstruction has an unknown scale: its distances do not directly correspond to real-world units. To recover metric scale, we use depth estimates $d_f^{\mathrm{M}}$ from frozen MoGe-2 as a reference. After resizing the MoGe-2~\citep{wang2026moge} depth maps to match the VGGT-Omega resolution, we estimate a single scale factor $s$ that minimizes the squared differences between the scaled VGGT-Omega depths and the MoGe-2 depths:
\begin{equation}
    s =
    \frac{\sum_{(f,p)\in\Omega}
    d_f^{\mathrm{V}}(p)d_f^{\mathrm{M}}(p)}
    {\sum_{(f,p)\in\Omega}
    \left(d_f^{\mathrm{V}}(p)\right)^2},
    \label{eq:metric_scale}
\end{equation}
where $\Omega$ contains valid, high-confidence pixel correspondences between the two depth predictions across frames. We apply $s$ to both the predicted depths and camera translations:
\begin{equation}
    \tilde d_f(p)=s d_f^{\mathrm{V}}(p),
    \qquad
    \tilde t_f=s t_f.
\end{equation}
Camera rotations and intrinsics remain unchanged.

\paragraph{2D-to-3D lifting.}
We then use the calibrated geometry to map each grounded bounding box into 3D. Since a box may include background pixels, we slightly shrink it and retain only valid pixels whose depths are close to the median depth within the box. Each retained pixel is first back-projected into the camera coordinate system using its depth and camera intrinsics, then transformed into the shared world coordinate system:
\begin{equation}
    \tilde{\mathbf{X}}_f(p)
    =
    R_f \left(\tilde d_f(p) K_f^{-1}\bar p\right)
    + \tilde t_f,
    \label{eq:2d_to_3d}
\end{equation}
where $\bar p=(u,v,1)^{\top}$ denotes the homogeneous pixel coordinates. We combine these points across frames while retaining their source-frame indices. The resulting metric point cloud places the grounded observations in a common 3D space for subsequent cross-view association and spatial reasoning.

\subsection{Association}

After reconstruction, observations from different frames are placed in the same 3D coordinate system, but it remains unclear which observations correspond to the same physical instance. This problem is further complicated by imperfect 2D grounding: bounding boxes predicted by the Video MLLM can be loose and include substantial background regions. Directly lifting and aggregating all points inside these boxes therefore introduces background points that vary across views and can make cross-view matching unreliable.

To obtain more robust geometric evidence, we partition the fused point cloud into geometrically coherent \emph{superpoints} following \cite{robert2023efficient}. Superpoints group nearby points with similar local geometry, allowing object surfaces to be separated from surrounding background structures. We construct a $k$-nearest-neighbor graph and apply cut-pursuit optimization~\citep{landrieu2017cut} using point coordinates and local geometric features such as planarity, surface orientation, and curvature.

For each grounded observation, its filtered 3D points vote for the superpoints they occupy. This converts a noisy set of box-level points into a compact set of geometric regions. Weak votes are discarded, and a one-hop dilation is applied over adjacent superpoints to connect partial observations of the same surface. To further suppress common background regions, we downweight superpoints that are shared by many observations. Let $c_{i,s}$ denote the vote count of observation $i$ for superpoint $s$, $\mathrm{df}(s)$ the number of observations voting for $s$, and $N$ the total number of observations:
\[
w_{i,s}
=
c_{i,s}
\left(
\log\frac{1+N}{1+\mathrm{df}(s)}
+1
\right).
\]

We then associate observations from different frames using their weighted superpoint evidence. Two observations are matched when their vote vectors have high cosine similarity and their 3D centers satisfy an extent-aware distance constraint. Compatible matches are merged using union-find, while merges causing excessive center drift are rejected to avoid connecting nearby but distinct instances.

For each query entity, the largest consistent component defines a view-consistent track, 3D anchor, and spatial footprint. Competing separated components are marked as conflicts, while insufficient superpoint evidence falls back to a robust bounding-box depth center marked as unverified. All association decisions rely on reconstructed geometry and fixed rules, without a learned cross-view matcher or benchmark-specific association training.

\subsection{Reasoning}

After reconstruction and association, GeoLift provides a view-consistent 3D map together with the original video to the frozen Video MLLM. The Video MLLM acts as the reasoning controller: it integrates geometric hints with visual evidence and selects appropriate tools from a geometric toolbox according to the spatial query.

The toolbox supports explicit spatial operations in the shared 3D space, such as computing relative directions and distances. Route planning is one representative example, where the model invokes occupancy-grid construction and A* search~\citep{hart1968formal} to connect successive landmarks and derive the required turns.

For queries that cannot be fully resolved by deterministic geometry, the reconstructed map provides metric anchors, object extents, distances, or floor footprints as additional hints to the Video MLLM. A reliability gate checks the consistency of the geometric evidence; when it is unreliable, the model relies more heavily on the original visual evidence or the direct Video MLLM prediction.

\section{Experiments}
\begin{table*}[t]
    \centering
    \setlength{\tabcolsep}{2.5pt}
    \caption{\textbf{Comparative results on full VSI-Bench}~\citep{2024Thinking}. The numerical and multiple-choice columns report MRA and exact-match accuracy, respectively. Training-free means no benchmark-specific updates; $^{\dagger}$ marks quoted results. Direct Qwen3-VL-4B and GeoLift use the same frozen backbone and matched evaluation protocol. All result values are percentages rounded to one decimal place.}
    \label{tab:vsi_main}
    \resizebox{\textwidth}{!}{%
    \begin{tabular}{l c cccc cccc}
        \toprule
        \multirow{2}{*}{\textbf{Methods}} &
        \multirow{2}{*}{\textbf{Overall}} &
        \multicolumn{4}{c}{\textbf{Numerical Questions}} &
        \multicolumn{4}{c}{\textbf{Multiple-Choice Questions}} \\
        \cmidrule(lr){3-6} \cmidrule(lr){7-10}
        & & Obj. Cnt. & Abs. Dist. & Obj. Size & Room Size &
        Rel. Dist. & Rel. Dir. & Route Plan & Appr. Order \\
        \midrule

        \multicolumn{10}{l}{\textit{General-Purpose MLLMs}} \\

        GPT-4o$^{\dagger}$~\citep{2024Thinking}
        & 34.0 & 46.2 & 5.3 & 43.8 & 38.2
        & 37.0 & 41.3 & 31.5 & 28.5 \\

Gemini-1.5 Flash$^{\dagger}$~\citep{2024Thinking}
        & 42.1 & 49.8 & 30.8 & 53.5 & 54.4
        & 37.7 & 41.0 & 31.5 & 37.8 \\

        Gemini-1.5 Pro$^{\dagger}$~\citep{2024Thinking}
        & 45.4 & 56.2 & 30.9 & 64.1 & 43.6
        & 51.3 & 46.3 & 36.0 & 34.6 \\

        \midrule

        InternVL2-8B$^{\dagger}$~\citep{2024Thinking}
        & 37.5 & 31.3 & 29.0 & 48.9 & 44.2
        & 38.0 & 33.4 & 28.9 & 46.4 \\

        InternVL3-78B$^{\dagger}$~\citep{Zhao2025SpaceMindCM}
        & 48.5 & 71.2 & 53.7 & 44.4 & 39.5
        & 55.9 & 39.5 & 28.9 & 55.4 \\

        Qwen2.5-VL-7B$^{\dagger}$~\citep{Zhao2025SpaceMindCM}
        & 33.0 & 40.9 & 14.8 & 43.4 & 10.7
        & 38.6 & 38.5 & 33.0 & 29.8 \\

        LLaVA-OneVision-72B$^{\dagger}$~\citep{Zhao2025SpaceMindCM}
        & 40.2 & 43.5 & 23.9 & 57.6 & 37.5
        & 42.5 & 39.9 & 32.5 & 44.6 \\

        LLaVA-NeXT-Video-72B$^{\dagger}$~\citep{Zhao2025SpaceMindCM}
        & 40.9 & 48.9 & 22.8 & 57.4 & 35.3
        & 42.4 & 36.7 & 35.0 & 48.6 \\

        Qwen3-VL-8B-Instruct$^{\dagger}$~\citep{Huang2026ConsiSpaceLG}
        & 57.4 & 66.7 & 38.9 & 74.1 & 53.5
        & 59.0 & 65.4 & 32.0 & 69.4 \\

        Qwen3-VL-4B-Instruct~\citep{Bai2025Qwen3VL}
        & 55.4 & 65.9 & 43.1 & 73.9 & 61.0
        & 53.1 & 46.1 & 30.9 & 69.1 \\

        \midrule
        \multicolumn{10}{l}{\textit{Training-Based Spatial MLLMs}} \\

        VG-LLM-4B$^{\dagger}$~\citep{Zheng2025LearningFV}
        & 47.3 & 66.0 & 37.8 & 55.2 & 59.2
        & 44.6 & 45.6 & 33.5 & 36.4 \\

        VG-LLM-8B$^{\dagger}$~\citep{Zheng2025LearningFV}
        & 50.7 & 67.9 & 37.7 & 58.6 & 62.0
        & 46.6 & 40.7 & 32.4 & 59.2 \\

        Spatial-MLLM$^{\dagger}$~\citep{Wu2025SpatialMLLMBM}
        & 48.4 & 65.3 & 34.8 & 63.1 & 45.1
        & 41.3 & 46.2 & 33.5 & 46.3 \\

        Cambrian-S-7B$^{\dagger}$~\citep{Yang2026CambrianS}
        & 67.5 & 73.2 & 50.5 & 74.9 & 72.2
        & 71.1 & 76.2 & 41.8 & 80.1 \\

        VLM-3R$^{\dagger}$~\citep{Fan2026VLM3R}
        & 60.9 & 70.2 & 49.4 & 69.2 & 67.1
        & 65.4 & 80.5 & 45.4 & 40.1 \\

        GS-Reasoner (w/d)$^{\dagger}$~\citep{Chen2026GSReasoner}
        & 64.7 & 69.1 & 61.9 & 70.0 & 65.7
        & 65.4 & 88.9 & 44.3 & 52.3 \\

        SpaceMind$^{\dagger}$~\citep{Zhao2025SpaceMindCM}
        & 69.6 & 73.3 & 61.4 & 77.3 & 74.2
        & 67.2 & 88.4 & 44.3 & 70.6 \\

        EgoMind (Qwen2.5-VL-7B)$^{\dagger}$~\citep{Chen2026EgoMind}
        & 50.2 & 55.4 & 37.9 & 67.1 & 40.4
        & 44.1 & 47.2 & 32.0 & 58.4 \\

        SpatioLM (SenseNovaSI-8B)$^{\dagger}$~\citep{Wu2026SpatioLM}
        & 71.6 & 81.9 & 55.7 & 75.2 & 71.5
        & 78.2 & 74.4 & 47.9 & 87.8 \\

        ConsiSpace (UC-SSRL)$^{\dagger}$~\citep{Huang2026ConsiSpaceLG}
        & 76.6 & 77.9 & 78.9 & 82.2 & 76.3
        & 72.1 & 90.1 & 54.7 & 79.9 \\

        SpaR3D-MoE$^{\dagger}$~\citep{Feng2026SpaR3DMoEA3}
        & 63.5 & 71.3 & 48.0 & 72.8 & 69.6
        & 58.7 & 70.1 & 44.0 & 73.5 \\

        VoxAnchor-4B$^{\dagger}$~\citep{Li2026VoxAnchor}
        & 66.8 & 66.3 & 52.6 & 72.5 & 71.0
        & 67.3 & 80.0 & 46.4 & 78.0 \\

        \midrule
        \multicolumn{10}{l}{\textit{Training-Free}} \\

        VideoAnchor$^{\dagger}$ (InternVL2-8B)~\citep{Wang2026VideoAnchor}
        & 37.8 & 39.3 & 30.3 & 50.5 & 38.8
        & 38.6 & 31.0 & 36.1 & 37.8 \\

TRACE$^{\dagger}$ (Qwen2.5-VL-72B-Instruct)~\citep{Hua2026TRACE}
        & 39.4 & 22.1 & 28.0 & 60.0 & 39.0
        & 40.9 & 37.4 & 32.0 & 42.6 \\

        Reason, Then Re-reason$^{\dagger}$ (Qwen3-VL-4B)~\citep{Ma2026ReReason}
        & 36.5 & 21.1 & 26.7 & 50.2 & 35.8
        & 37.2 & 43.1 & 28.9 & 34.1 \\

\textbf{Ours (Qwen3-VL-4B)}
        & \textbf{64.1} & \textbf{68.0} & \textbf{49.4}
        & \textbf{74.4} & \textbf{65.9}
        & \textbf{57.8} & \textbf{74.4} & \textbf{53.6} & \textbf{69.1} \\

        \bottomrule
    \end{tabular}%
    }
\end{table*}

\subsection{Experimental Setup}

We evaluate GeoLift on the official full splits of VSI-Bench~\citep{2024Thinking} and MindCube~\citep{yin2025mindcube}. VSI-Bench contains 5,130 questions from 288 egocentric videos; its overall score is the unweighted mean of eight category scores, using MRA for numerical tasks and exact-match accuracy for multiple-choice tasks. MindCube contains 21,154 sparse-multiview questions; we report Rotation, Among, and Around accuracies together with the pooled micro-average over all questions.

All neural components remain frozen, with no benchmark-specific updates or spatial-QA supervision. Published baselines retain their reported protocols; the direct Qwen3-VL-4B baseline and a matched no-superpoint variant provide controlled comparisons.

\subsection{Main Results on VSI-Bench}
Table~\ref{tab:vsi_main} shows that GeoLift produces task-dependent rather than uniform improvements. Relative to direct Qwen3-VL-4B-Instruct inference, GeoLift raises the VSI-Bench category macro-average from 55.4 to 64.1, with the largest gains in relative direction (+28.3 percentage points) and route planning (+22.7 percentage points). These gains are concentrated in tasks that require integrating evidence across multiple views, maintaining a persistent global spatial state, and composing several spatial relations, particularly in long-horizon route planning. By contrast, object counting and object-size estimation are comparatively local: an informative frame or a small number of frames may be sufficient, and size judgments may additionally draw on local visual cues and category-level priors acquired during pretraining. Their weaker dependence on cross-view integration limits the marginal benefit of explicit 3D reconstruction. GeoLift also exceeds several listed training-based spatial baselines, including VG-LLM-8B (50.7), Spatial-MLLM (48.4), VLM-3R (60.9), and SpaR3D-MoE (63.5), as well as the direct Qwen3-VL-8B-Instruct baseline (57.4) and the reported training-free baselines TRACE (39.4), VideoAnchor (37.8), and Reason, Then Re-reason (36.5). 

\subsection{Results on MindCube}
To assess transfer beyond egocentric video, Table~\ref{tab:mindcube} reports GeoLift's performance on MindCube, which uses sparse multiview observations. We use the same frozen semantic and geometric components as on VSI-Bench.

\begin{table}[t]
    \centering
    \caption{\textbf{Results on the full MindCube benchmark}~\citep{yin2025mindcube}.
    We report accuracy for the Rotation, Among, and Around settings.
    The Around setting includes the AroundNew subset. The overall score is
    the micro-averaged accuracy over all 21,154 questions.  GeoLift is
    training-free: the Qwen3-VL-4B variant uses Qwen3-VL-4B for grounding
    and answering. Bold numbers
    indicate the best value in each column. All values are percentages
    rounded to one decimal place.}
    \label{tab:mindcube}
    \setlength{\tabcolsep}{5pt}
    \begin{tabular}{lrrrr}
        \toprule
        \textbf{Method} & \textbf{Overall (micro)} & \textbf{Rotation} &
        \textbf{Among} & \textbf{Around} \\
        \midrule

        \multicolumn{5}{l}{\textit{Baseline}} \\
        Chance
        & 32.4 & 36.4 & 32.3 & 30.7 \\

        \midrule
        \multicolumn{5}{l}{\textit{Open-Weight Multi-Image Models}} \\
        LLaVA-OneVision-7B
        & 47.4 & 36.5 & 48.4 & 44.1 \\
        Qwen2.5-VL-7B-Instruct
        & 29.3 & 38.8 & 29.5 & 21.4 \\
        Gemma-3-12B-it
        & 46.7 & 38.4 & 48.4 & 34.6 \\
        Mantis-8B (SigLip)
        & 41.1 & 37.7 & 40.2 & 51.0 \\
        DeepSeek-VL2-Small
        & 47.6 & 37.0 & 50.4 & 26.9 \\
        Qwen3-VL-4B 
        & 23.4 & 32.0 & 21.9 & 33.2 \\

        \midrule
        \multicolumn{5}{l}{\textit{Proprietary Models}} \\
        GPT-5
        & 47.6 & \textbf{93.3} & 34.2 & 41.6 \\
        Claude-4-Sonnet
        & 44.8 & 48.4 & 44.2 & 47.6 \\
        GPT-5.6-Sol 
        & 54.1 & 67.1 & 52.6 & 61.5 \\

        \midrule
        \multicolumn{5}{l}{\textit{Spatial Models}} \\
        RoboBrain
        & 37.4 & 35.8 & 38.3 & 29.5 \\
        SpaceMantis
        & 22.8 & 37.7 & 21.3 & 29.3 \\
        Space-Qwen
        & 33.3 & 38.0 & 33.7 & 26.3 \\

        \midrule
        \textbf{Ours (Qwen3-VL-4B)}
        & \textbf{59.8} & 56.2 & \textbf{58.5} & \textbf{74.6} \\

        \bottomrule
    \end{tabular}
\end{table}

On the full MindCube benchmark, GeoLift achieves 59.8\% micro-averaged accuracy, improving over direct Qwen3-VL-4B inference (23.4\%) by 36.4 percentage points. Performance is highest on Around (74.6\%), followed by Among (58.5\%) and Rotation (56.2\%), a pattern consistent with the value of organizing sparse observations into a shared spatial representation. GeoLift also exceeds the GPT-5.6-Sol baseline in overall accuracy (54.1\%) and on Around (61.5\%). GPT-5 nevertheless retains the highest Rotation score (93.3\%). Together, these results complement VSI-Bench by supporting the usefulness of explicit spatial organization beyond egocentric video.

\subsection{Analysis and Ablations}
Table~\ref{tab:ablation} isolates the contributions of metric geometry and superpoint association. Introducing the VGGT-Omega/MoGe-2 geometry branch raises the category macro-average from 55.4 to 62.8, with particularly pronounced improvements in relative direction (+26.0 points) and route planning (+16.0 points). Superpoint association further increases the macro-average to 64.1, contributing the largest additional gain to route planning (+6.7 points), followed by relative direction (+2.3 points), with further improvements in absolute-distance and room-size estimation. The concentration of these gains in tasks that require integrating observations across views and composing spatial relations is consistent with GeoLift's intended role: bridging frame-level visual perception and scene-level spatial reasoning through a shared spatial reference. 

\begin{table*}[t]
    \centering
    \setlength{\tabcolsep}{2.5pt}
    \caption{\textbf{Ablation of metric geometry and superpoint association on VSI-Bench.} All rows use the frozen Qwen3-VL-4B Video MLLM. A checkmark indicates that the component is enabled, while a cross indicates that it is disabled. The first row is direct Qwen3-VL-4B inference without VGGT-Omega, MoGe-2, or superpoint association. Geometric variants use the fixed 64-frame protocol. Scores are percentages.}
    \label{tab:ablation}
    \resizebox{\textwidth}{!}{%
    \begin{tabular}{c c c cccc cccc}
        \toprule
        \multirow{2}{*}{\textbf{VGGT+MoGe}} &
        \multirow{2}{*}{\textbf{SP Association}} &
        \multirow{2}{*}{\textbf{Overall}} &
        \multicolumn{4}{c}{\textbf{Numerical Questions}} &
        \multicolumn{4}{c}{\textbf{Multiple-Choice Questions}}
        \tabularnewline
        \cmidrule(lr){4-7}
        \cmidrule(lr){8-11}
        & & &
        \textbf{Obj. Cnt.} &
        \textbf{Abs. Dist.} &
        \textbf{Obj. Size} &
        \textbf{Room Size} &
        \textbf{Rel. Dist.} &
        \textbf{Rel. Dir.} &
        \textbf{Route Plan.} &
        \textbf{Appr. Order}
        \tabularnewline
        \midrule

        $\times$ & $\times$
        & 55.4
        & 65.9 & 43.1 & 73.9 & 61.0
        & 53.1 & 46.1 & 30.9 & 69.1
        \tabularnewline

        $\checkmark$ & $\times$
        & 62.8
        & 68.0 & 48.4 & 74.4 & 65.5
        & 57.8 & 72.1 & 46.9 & 69.1
        \tabularnewline

        $\checkmark$ & $\checkmark$
        & \textbf{64.1}
        & \textbf{68.0} & \textbf{49.4} & \textbf{74.4} & \textbf{65.9}
        & \textbf{57.8} & \textbf{74.4} & \textbf{53.6} & \textbf{69.1}
        \tabularnewline

        \bottomrule
    \end{tabular}%
    }
\end{table*}

\section{Limitations}
GeoLift's training-free design relies on frozen visual and geometric modules, making performance sensitive to 2D grounding and the reconstructed spatial reference. The Video MLLM determines entity coverage and localization, while VGGT-Omega supplies camera poses and depth and MoGe-2 calibrates metric scale; residual errors can propagate to spatial relations, metric estimates, and route planning. Reliability checks and fallback to direct Video MLLM answers mitigate uncertain geometry but cannot recover missing or ambiguous evidence. Future work could complement this training-free design with learning-based adaptation of grounding, reconstruction, metric calibration, and spatial reasoning to improve robustness and performance across task categories while preserving the explicit 3D representation.

\section{Conclusion}
Current Video MLLMs can recognize and localize query-relevant objects in individual frames but remain less reliable when integrating spatial evidence across viewpoints and multiple steps. We introduced GeoLift, a training-free framework that complements a frozen Video MLLM with a persistent metric 3D representation; for video inputs, geometric superpoints associate observations across views, while explicit spatial operators compute relations and routes. With the same frozen Qwen3-VL-4B backbone, GeoLift raises the VSI-Bench category macro-average from 55.4\% to 64.1\% and MindCube overall accuracy from 23.4\% to 59.8\%, with the largest improvements in relative direction and route planning. Although the gains are task-dependent, these results indicate that pairing pretrained visual grounding with an explicit 3D spatial state broadens the range of tasks that current Video MLLMs can address reliably, providing a practical path toward stronger cross-view and multistep reasoning without benchmark-specific parameter updates.

\bibliography{references}
\bibliographystyle{plainnat}

\end{document}